\documentclass[letterpaper, 10 pt, conference]{ieeeconf}  

\IEEEoverridecommandlockouts                              

\usepackage{cite}
\usepackage{mdwmath}
\usepackage{mdwtab}
\usepackage[caption=false]{subfig}
\usepackage{url}

\usepackage{graphics} 
\usepackage{epsfig} 
\usepackage{mathptmx} 
\DeclareMathAlphabet{\mathcal}{OMS}{cmsy}{m}{n}
\usepackage{times} 
\usepackage{amsmath} 
\usepackage{amssymb}  

\usepackage{hyperref}

\usepackage[utf8]{inputenc}
\usepackage{cleveref}
\usepackage{booktabs}
\usepackage{adjustbox}
\usepackage{threeparttable}
\usepackage{pifont}
\usepackage{siunitx}
\usepackage{placeins}
\usepackage{microtype}
\usepackage{balance}
\usepackage{paralist}
\renewcommand{\baselinestretch}{0.97}
\usepackage[nolist,nohyperlinks]{acronym}
\newacro{CNN}{Convolutional Neural Network}
\newacro{DNN}{Deep Neural Network}
\newacro{GPS}{Global Positioning System}
\newacro{GNSS}{Global Navigation Satellite System}
\newacro{NLOS}{non-line-of-sight}
\newacro{ADAS}{Advanced Driver Assistance Systems}
\newacro{LIDAR}[LiDAR]{Light Detection And Ranging}
\newacro{HD map}{High Definition map}
\newacro{EV}{Embedding Vector}
\newacro{SLAM}{Simultaneouos Localization And Mapping}
\newacro{MLP}{MultiLayer Perceptron}
\newacro{IMU}{Inertial Measurement Unit}
\newacro{ML}{Machine Learning}
\newacro{SfM}{Structure from Motion}
\newacro{PnP}{Perspective-n-Points}
\newacro{ASPP}{Atrous Spatial Pyramid Pooling}
\newacro{RANSAC}{RANdom SAmple Consensus}
\newacro{CV}{Computer Vision}

\def\eg{\emph{e.g., }}
\def\ie{\emph{i.e., }}
\def\wrt{\emph{w.r.t. }}
\def\cmrnet2{CMRNet++}

\title{\LARGE \bf
\cmrnet2: Map and Camera Agnostic\\Monocular Visual Localization in LiDAR Maps
}

\author{Daniele Cattaneo$^{1,2}$, Domenico Giorgio Sorrenti$^{2}$, Abhinav Valada$^{1}$
\thanks{$^{1}$ Dep. of Computer Science, University of Freiburg, Germany}%
\thanks{$^{2}$ Dep. Informatica, Sistem. e Comun., Università di Milano - Bicocca, Italy}%
}

\usepackage{tikz}
\usepackage{textcomp}
\usepackage{lipsum}

\begin{document}

\maketitle

\thispagestyle{empty}
\pagestyle{empty}

\begin{abstract}
Localization is a critically essential and crucial enabler of autonomous robots. While deep learning has made significant strides in many computer vision tasks, it is still yet to make a sizeable impact on improving capabilities of metric visual localization. One of the major hindrances has been the inability of existing \ac{CNN}-based pose regression methods to generalize to previously unseen places. Our recently introduced CMRNet effectively addresses this limitation by enabling map independent monocular localization in LiDAR-maps. In this paper, we now take it a step further by introducing \cmrnet2, which is a significantly more robust model that not only generalizes to new places effectively, but is also independent of the camera parameters. We enable this capability by combining deep learning with geometric techniques, and by moving the metric reasoning outside the learning process. In this way, the weights of the network are not tied to a specific camera. Extensive evaluations of \cmrnet2 on three challenging autonomous driving datasets, \ie KITTI, Argoverse, and Lyft5, show that \cmrnet2 outperforms CMRNet as well as other baselines by a large margin. More importantly, for the first-time, we demonstrate the ability of a deep learning approach to accurately localize without any retraining or fine-tuning in a completely new environment and independent of the camera parameters.\looseness=-1

\end{abstract}

\section{Introduction}\label{sec:introduction}
Autonomous mobile robots, such as self-driving cars require accurate localization to safely navigate. Although \acp{GNSS} provides global positioning, its accuracy and reliability is not adequate for robot navigation. For example, in urban environments, buildings often block or reflect satellites signals, causing \ac{NLOS} and multipath issues. In order to alleviate this problem, localization methods that exploit sensors on the robot are used to improve precision and robustness.

A wide range of methods have been proposed to tackle the localization task, using a variety of onboard sensors. While LiDAR-based approaches~\cite{kummerle2015autonomous,behley2018efficient} typically achieve sufficiently accurate localization, their adoption is primarily hindered due to the associated high cost. On the other hand, camera-based methods~\cite{sattler2018benchmarking, sattler2016efficient, boniardi2019robot} are more promising for widespread adoption in autonomous vehicles as they are significantly inexpensive. Although historically the performance of camera-based approaches have been subpar compared to LiDAR-based methods, recent advances in computer vision and machine learning have substantially narrowed this gap. Some of these methods employ \acp{CNN}~\cite{Kendall_2015_ICCV,8458420,valada2018deep,Brachmann_2018_CVPR,valada2018incorporating} and random forests~\cite{Shotton_2013_CVPR,cavallari2017fly} to directly regress the pose of the camera given a single image. Although these methods have achieved remarkable results in indoor environments, their performance has been significantly limited in large-scale outdoor environments~\cite{sattler2019understanding}. Moreover, they can only be employed in locations where these models have been previously trained on.

In the last decade, map providers have been developing the next generation HD maps tailored for the automotive domain. These maps include accurate geometric reconstructions of road scenes in the form of point clouds, most often generated from LiDARs. This factor has motivated researchers to develop methods to localize a camera inside LiDAR-maps. Localization can typically be performed by reconstructing the three-dimensional geometry of the scene from a camera, and then matching this reconstruction with the map~\cite{Caselitz_2016, sun2019scale}, or by matching in the image plane~\cite{Wolcott_2014,neubert2017sampling,Cattaneo_2019}.

In this paper, we present our \cmrnet2 approach for camera to LiDAR-map registration. We build upon our previously proposed CMRNet~\cite{Cattaneo_2019} model, which was inspired by camera-to-LiDAR pose calibration techniques~\cite{Schneider_2017}. CMRNet localizes independent of the map, and we now further improve it to also be independent of the camera intrinsics. Unlike existing state-of-the-art \ac{CNN}-based approaches for pose regression~\cite{Kendall_2015_ICCV,Brachmann_2018_CVPR,8458420}, CMRNet does not learn the map, instead it learns to match images to a pre-existing map. Consequently, CMRNet can be used in any environment for which a LiDAR-map is available. However, since the output of CMRNet is metric (a 6-DoF rigid body transformation from an initial pose), the weights of the network are tied to the intrinsic parameters of the camera used for collecting the training data. In this work, we mitigate this problem by decoupling the localization, by first employing a pixel to 3D point matching step, followed by a pose regression step. While the network is independent of the intrinsic parameters of the camera, they are still required for the second step.
We evaluate our model on KITTI~\cite{Geiger2013IJRR}, Agroverse~\cite{Argoverse}, and Lyft5~\cite{lyft2019} datasets and demonstrate that our approach exceeds state-of-the-art methods while being agnostic to map and camera parameters. A live demo and videos showing qualitative results of our approach on each of these datasets are available at \href{https://rl.uni-freiburg.de/research/vloc-in-lidar}{\color{red}http://rl.uni-freiburg.de/research/vloc-in-lidar}.


\section{Technical Approach}\label{sec:proposed-approach}

\begin{figure*}
\centering
    \includegraphics[width=.85\textwidth,trim={0 7.26cm 0.7cm 0},clip]{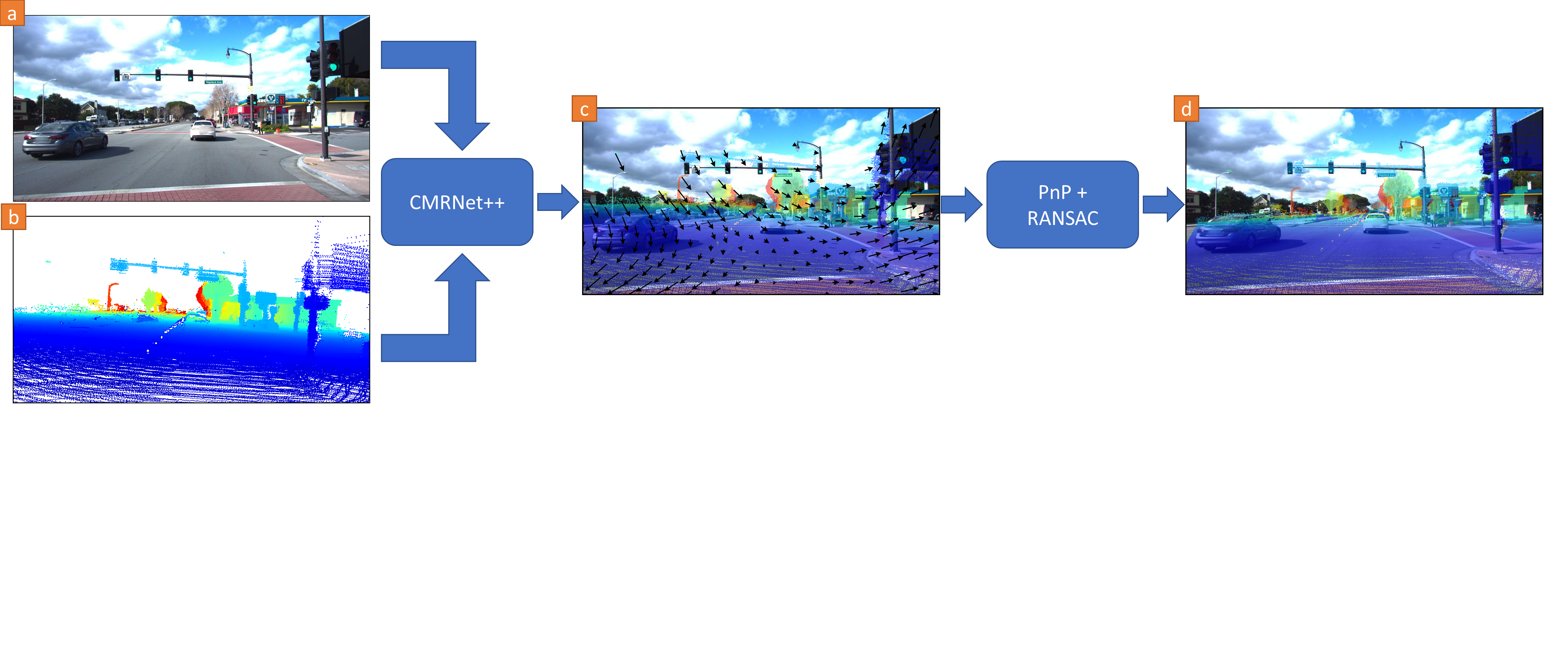}
    \caption{Outline of the proposed approach. (a) The input RGB image and (b) the LiDAR-image are fed to our \cmrnet2, which predicts (c) pixel displacements between the two inputs. (d) The predicted matches are used to localize the camera using a PnP+RANSAC scheme.}
    \label{fig:pipeline}
    \vspace{-0.2cm}
\end{figure*}

We extend CMRNet by decoupling the localization into two steps: pixel to 3D point matching, followed by pose regression. In the first step, the \ac{CNN} only focuses on matching at the pixel-level instead of metric basis, which makes the network independent of the intrinsic parameters of the camera. These parameters are instead employed in the second step, where traditional computer vision methods are exploited to estimate the pose of the camera, given the matches from the first step. Consequently, after training, \cmrnet2 can also be used with different cameras and maps from those used while training. An outline of our proposed \cmrnet2 pipeline is depicted in \Cref{fig:pipeline}.

\subsection{Matching Step}
For the matching step, we generate a synthesized depth image, which we refer to as a LiDAR-image, by projecting the map into a virtual image plane placed at $H_{init}$, a rough pose estimate obtained, \eg from a \ac{GNSS}. The projection uses the camera intrinsics. \Cref{fig:pipeline}(b) shows an example of a LiDAR-image. 
To deal with occlusions in point clouds, we employ a z-buffer technique followed by an occlusion estimation filter, see~\cite{Cattaneo_2019} for more details. Once the inputs to the network (camera and LiDAR images) have been obtained, for every 3D point in the LiDAR-image, \cmrnet2 estimates the pixel of the RGB image that represents the same world point. 

The architecture of \cmrnet2 is based on PWC-Net~\cite{Sun_2018_CVPR}, which was introduced for optical flow estimation between two consecutive RGB frames. Unlike PWC-Net, \cmrnet2 does not share weights between the two feature pyramid extractors, as the inputs to \cmrnet2 are inherently different (RGB and LiDAR images). Moreover, since the LiDAR-image only has one channel (\ie depth), we change the number of input channels of the first convolutional layer in the feature extractor, from three to one. The output of \cmrnet2 is a dense feature map, which is $1/4$-th the input resolution and consists of two channels that represent, for every pixel in the LiDAR-image, the displacement ($u$ and $v$) of the pixel in the RGB image from the same world point. A visual representation of this pixel displacement is depicted in \Cref{fig:pipeline}~(c).

In order to train \cmrnet2, we first need to generate the ground truth pixel displacement $\Delta P$ of the LiDAR-image \wrt RGB image. To accomplish this, we first compute the coordinates of the map's points in the $H_{init}$ reference frame using \Cref{eq:frame_init} and the pixel position of their projection in the LiDAR-image exploiting the intrinsic matrix $K$ of the camera from \Cref{eq:proj_init}.
\begin{gather}
    [
    \mathbf{x}^{init} \
    \mathbf{y}^{init} \
    \mathbf{z}^{init} \
    \mathbf{1}
    ]^\intercal
     =  H^{init}_{map} \cdot 
    [ 
    \mathbf{x}^{map} \
    \mathbf{y}^{map} \
    \mathbf{z}^{map} \
    \mathbf{1}
    ]^\intercal,    \label{eq:frame_init}\\
    [ 
    \mathbf{u}^{init} \
    \mathbf{v}^{init} \
    \mathbf{1}
    ]^\intercal
     = K \cdot 
    [ 
    \mathbf{x}^{init} \
    \mathbf{y}^{init} \
    \mathbf{z}^{init} \
    \mathbf{1}
    ]^\intercal.  \label{eq:proj_init}
\end{gather}

We keep track of indices of valid points in an array $\mathbf{VI}$. This is done by excluding indices of points whose projection lies behind or outside the image plane, as well as points marked occluded by the occlusion estimation filter. Subsequently, we generate the sparse LiDAR-image $\mathcal{D}$ and project the points of the map into a virtual image plane placed at the ground truth pose $H_{GT}$. We then store the pixels' position of these projections as
\begin{gather}
\mathcal{D}_{\mathbf{u}^{init}_i, \mathbf{v}^{init}_i} = \mathbf{z}^{init}_i,\  i \in \mathbf{VI}, \label{eq:lidar_img}\\
[ 
    \mathbf{u}^{GT} \
    \mathbf{v}^{GT} \
    \mathbf{1}
    ]^\intercal
    = K \cdot H^{GT}_{map} \cdot 
    [ 
    \mathbf{x}^{map} \
    \mathbf{y}^{map} \
    \mathbf{z}^{map} \
    \mathbf{1}
    ]^\intercal. \label{eq:proj_GT}
\end{gather}

Finally, we compute the displacement ground truths $\Delta P$ by comparing the projections in the two image planes as
\begin{equation}
\label{eq:GT}
\Delta P_{\mathbf{u}^{init}_i, \mathbf{v}^{init}_i} = \left[ \mathbf{u}^{GT}_i-\mathbf{u}^{init}_i, \mathbf{v}^{GT}_i-\mathbf{v}^{init}_i \right], \ i \in \mathbf{VI}.
\end{equation}

For every pixel $[u, v]$ without an associated 3D point, we set $\mathcal{D}_{u, v}=0$ and $\Delta P_{u, v}=[0,0]$. Moreover, we generate a mask of valid pixels as $mask_{u,v} = 1 \ if \ \mathcal{D}_{u,v}>0, \ 0\ \text{otherwise}$. We use a loss function that minimizes the sum of the regression component $\mathcal{L}_{reg}$ and the smoothness component $\mathcal{L}_{smooth}$, to train our network. The regression loss defined in \Cref{eq:loss_reg} penalizes pixel displacements predicted by the network $\widehat{\Delta P}$  that differs from their respective ground truth displacements $\Delta P$ on valid pixels. While the smoothness loss $\mathcal{L}_{smooth}$ enforces the displacement of pixels without a ground truth to be similar to the ones in the neighboring pixels.
\begin{alignat}{2} 
&\mathcal{L}_{reg} = \frac{\sum_{u,v} \ \| \widehat{\Delta P}_{u,v} - \Delta P_{u, v} \| \cdot mask_{u, v}}{\sum_{u,v} \ mask_{u, v}}\label{eq:loss_reg} \\
\begin{split}\label{eq:loss_smooth}
&D_{smoooth}(u,v) = \begin{aligned}[t] 
&\rho (\widehat{\Delta P}_{u, v}-\widehat{\Delta P}_{u+1, v}) \\
& + \rho (\widehat{\Delta P}_{u, v}-\widehat{\Delta P}_{u, v+1})
\end{aligned}
\end{split}\\
&\mathcal{L}_{smooth} = \frac{\sum_{u,v} \ D_{smoooth}(u,v)
\cdot \left( 1 - mask_{u, v} \right)}{\sum_{u,v} \ \left( 1-mask_{u, v} \right) }
\end{alignat}
where $\rho$ is the generalized Charbonnier function $\rho (x) = (x^2 + \epsilon^2)^\alpha, \epsilon=10^{-9}, \alpha=0.25$, as in~\cite{unsupervisedflow}.

\subsection{Localization Step}
\label{sec:localization}
Once \cmrnet2 has been trained, we have the map, \ie a set of 3D points $\mathbf{P}$ whose coordinates are known, altogether with their projection in the LiDAR-image $\mathcal{D}$ and a set $\mathbf{p}$ of matching points in the RGB image that is predicted by the \ac{CNN} given as
\begin{gather}
    \mathbf{P}_i = [\mathbf{x}_i, \mathbf{y}_i, \mathbf{z}_i], \ i \in \mathbf{VI}, \\
    \mathbf{p}_i = [\mathbf{u}^{init}_i, \mathbf{v}^{init}_i] + \widehat{\Delta P}_{\mathbf{u}^{init}_i,\mathbf{v}^{init}_i} , \ i \in \mathbf{VI}.
\end{gather}

Estimating the pose of the camera given a set of 2D-3D correspondences and the camera intrinsics is known as the \ac{PnP} problem. We use the EPnP algorithm~\cite{lepetit2009epnp} within a \ac{RANSAC} scheme~\cite{fischler1981random} to solve this, with a maximum of 1000 iterations and an inlier threshold value of 2 pixels.

\subsection{Iterative Refinement}
\label{sec:iterative}
Similar to CMRNet, we employ an iterative refinement technique where we train different instances of \cmrnet2, each specialized in handling different initial error ranges. During inference, we feed the RGB and LiDAR-image to the network trained with the highest error range, and we generate a new LiDAR-image by projecting the map in the predicted pose. The latter is then fed to the second instance of \cmrnet2 that is trained with a lower error range. This process can be repeated multiple times, iteratively improving the estimated localization accuracy.

\subsection{Training Details}
We train each instance of \cmrnet2 from scratch for $300$ epochs with a batch size of $40$ using two NVIDIA Tesla P100. The weights of the network were updated with the ADAM optimizer with an initial learning rate of $1.5 \cdot 10^{-4}$ and a weight decay of $5\cdot 10^{-6}$. We halved the learning rate after $20$ and $40$ epochs.

\section{Experimental Results}\label{sec:experimental-results}

\subsection{Datasets}
\label{sec:datasets}
To evaluate the localization performance of our \cmrnet2 and to assess its generalization ability, we use three diverse autonomous driving datasets that cover different countries, different sensors, and different traffic conditions.

\subsubsection{KITTI}
The KITTI dataset~\cite{Geiger2013IJRR} was recorded around the city of Karlsruhe, Germany.
We use the \textit{left} camera images from the odometry sequences 03, 05, 06, 07, 08 and 09 as the training set (total of \num{11426} frames), and the sequence 00 as the validation set (\num{4541} frames). 

\subsubsection{Argoverse}
The Argoverse dataset~\cite{Argoverse} was recorded in Miami and Pittsburgh.
We used the images from the central forward facing camera that provides $1920 \times 1200$ images at 30 fps.
The images from \textit{train1}, \textit{train2} and \textit{train3} splits of the ``3D tracking dataset'' were used as the training set (\num{36347} frames), and the \textit{train4} split as the validation set (\num{2741} frames).

\subsubsection{Lyft5}
The Lyft Level 5 AV dataset~\cite{lyft2019} was recorded in Palo Alto.
We use the $1224\times1024$ images recorded from the front camera from 10 selected urban scenes as validation set (\num{1250} frames).
We utilize the Lyft5 dataset to assess the generalization ability of our approach. We first train our \cmrnet2 on KITTI and Argoverse, and then evaluate it on the Lyft5 dataset to assess the localization ability without any retraining. Therefore, this dataset was not included in the training set.

\subsection{LiDAR-maps generation}
In order to generate LiDAR-maps for the three aforementioned datasets, we first aggregate single scans at their respective ground truth position, which is either provided by the dataset itself (Argoverse and Lyft5) or generated with a SLAM system~\cite{kummerle2015autonomous} (KITTI). We then downsample the resulting maps at a resolution of \SI{0.1}{\metre} using the Open3D library~\cite{Zhou2018}. Moreover, as we would like to have only static objects in the maps (\eg no pedestrians or cars), we remove dynamic objects by exploiting the 3D bounding boxes provided with Argoverse and Lyft5. Unfortunately, KITTI does not provide such bounding boxes for the odometry sequences, and therefore we could not remove dynamic objects for these sequences. In the future, we will leverage semantic segmentation techniques or 3D object detectors to remove them.

\begin{table*}
\centering
\begin{threeparttable}
\caption{Median localization error of \cmrnet2 on KITTI, Argoverse, and Lyft5 datasets.}
\label{tab:iterative}
\setlength\tabcolsep{4.2pt}
\begin{tabular*}{\linewidth}{lcc|ccc|ccc|ccc}
\toprule
 & \multicolumn{2}{c}{Training Max. Error Range}& \multicolumn{3}{c}{KITTI Localization Error}& \multicolumn{3}{c}{Argoverse Localization Error}& \multicolumn{3}{c}{Lyft5 Localization Error} \\ \cmidrule(lr){2-3} \cmidrule(lr){4-6} \cmidrule(lr){7-9} \cmidrule(lr){10-12}
            & Transl. [m] & Rot. [deg]  & Transl. [m] & Rot. [deg] & Fail [\%] & Transl. [m] & Rot. [deg] & Fail [\%] & Transl. [m] & Rot. [deg] & Fail [\%] \\ \midrule
Initial pose & - & - & $\approx1.97$ & $\approx9.83$ & - & $\approx1.97$ & $\approx9.83$ & - & $\approx1.97$ & $\approx9.83$ & - \\
Iteration 1 & [-2, +2]& [$-10$, $+10$]  &   0.55         &  $1.46$ & 2.18 &   0.80       &  $1.55$ &6.24&   1.32       &  $2.13$ & 10.56   \\
Iteration 2 & [-1, +1]& [$-2$, $+2$]    &   0.22         &  $0.77$ &-&   0.34       &  $0.58$ &-&   0.79       &  $1.28$ & -  \\
Iteration 3 & [-0.6, +0.6]& [$-1$, $+1$]&  \textbf{0.14} &  $\textbf{0.43}$  &-&  \textbf{0.25} &  $\textbf{0.45}$ &-&  \textbf{0.70} &  $\textbf{1.18}$ & -  \\ \bottomrule
\end{tabular*}
    \begin{tablenotes}[para,flushleft]
      \footnotesize      
      \item The initial poses were randomly sampled from a uniform distribution in [-2m, +2m], [-10,+10], the first line shows the corresponding median values. Note that \cmrnet2 was trained on KITTI and Argoverse, and only evaluated on Lyft5 to assess the generalization ability without any retraining.
    \end{tablenotes}
\end{threeparttable}
\vspace{-0.2cm}
\end{table*}

\begin{table}
\centering
\begin{threeparttable}
\caption{Comparison with state-of-the-art monocular approaches.}
\label{tab:comparison}
\setlength\tabcolsep{3.6pt}
\begin{tabular}{lcccccc}
\toprule
& \multicolumn{3}{c}{Translation [m]} & \multicolumn{3}{c}{Rotation [deg]} \\ \cmidrule(lr){2-4} \cmidrule(lr){5-7}
& Median &Mean & Std. Dev. & Median &Mean & Std. Dev. \\
\midrule
Caselitz~\cite{Caselitz_2016} &-&$0.30 $ & $\textbf{0.11}$ & -&$1.65$ & $0.91$ \\
CMRNet~\cite{Cattaneo_2019}   &0.27& $0.33 $ & $0.22$ & 1.07 & $1.07 $ & $0.77$ \\
\cmrnet2 & $\textbf{0.14}$ & $\textbf{0.21}$   & $0.30$ & $\textbf{0.43}$& $\textbf{0.52}$ & $\textbf{0.42}$ \\
\bottomrule

\end{tabular}
    \begin{tablenotes}[para,flushleft]
      \footnotesize      
      \item Localization errors comparison on the KITTI odometry sequence 00.
    \end{tablenotes}
\end{threeparttable}
\end{table}



\subsection{Training on multiple datasets}
We train \cmrnet2 by combining training samples from KITTI and Argoverse datasets. Training a \ac{CNN} on multiple diverse datasets creates certain challenges. First, the different cardinality of the two training datasets (\num{11426} and \num{36347}, respectively) might lead the network to perform better on one dataset than the other. To overcome this problem, we randomly sampled a subset of Argoverse at the beginning of every epoch to have the same number of samples as KITTI. As the subset is sampled every epoch, the network will eventually process every sample from the Argoverse dataset.

Moreover, the two datasets have cameras with very different field of view and resolution. To address this issue, we resize the images so to have the same resolution.
One straightforward way to accomplish this would be to just crop the Argoverse images. However, this would yield images with a very narrow field of view, making the matching between the RGB and LiDAR-image increasingly difficult. Therefore, we first downsample the Argoverse images to half the original resolution ($960\times600$), and then randomly crop both Argoverse and KITTI images to $960\times320$ pixels. We perform this random cropping at runtime during training in order to have different crop positions for the same image at different epochs.
We generate both the LiDAR-image $\mathcal{D}$ and the ground truth displacements $\Delta P$ at the original resolution, and then downsample and crop them accordingly. Moreover, we also halve the pixel displacements $\Delta P$ during the downsampling operation.

\subsection{Initial pose sampling and data augmentation}
We employ the iterative refinement approach presented in \Cref{sec:iterative} by training three instances of \cmrnet2. To simulate the initial pose $H_{init}$, we add uniform random noise to all components of the ground truth pose $H_{GT}$, independent for each sample. The range of the noise that we add to the first iteration is [$\pm 2$ m] for the translation and [$\pm$ \ang{10}] for the rotation. The maximum range for the second and third iteration are [$\pm 1$ m, $\pm$ \ang{2}] and [$\pm 0.6$ m, $\pm$ \ang{1}], respectively.

To improve the generalization ability of our approach, we employ a data augmentation scheme. First, we apply color jittering to the images by randomly changing the contrast, saturation, and brightness. Subsequently, we randomly horizontally mirror the images
, and we modify the camera calibration accordingly. We then randomly rotate the images in the range [\ang{-5}, \ang{+5}]. Finally, we transform the LiDAR point cloud to reflect these changes before generating the LiDAR-image.

To summarize, we train every instance of \cmrnet2 as follows:
\begin{compactitem}
    \item Randomly draw a subset of the Argoverse dataset and shuffle it with samples from KITTI
    \item For every batch:
\begin{compactitem}
    \item Apply data augmentation to the images $\mathcal{I}$ and modify the camera matrices and the point clouds accordingly
    \item Sample the initial poses $H_{init}$
    \item Generate the LiDAR-images $\mathcal{D}$, the displacement ground truths $\Delta P$ and the masks $mask$
    \item Downsample $I, \mathcal{D}, \Delta P$ and $mask$ for the Argoverse samples in the batch
    \item Crop $I, \mathcal{D}, \Delta P$ and $mask$ to the resolution $960\times320$.
    \item Feed the batch ($I, \mathcal{D}$) to \cmrnet2, compute the loss, and update the weights
\end{compactitem}
\item Repeat for $300$ epochs

\end{compactitem}

\subsection{Inference}
During inference, we process one image at a time, and, since the network architecture is fully convolutional, we do not resize the images of different datasets to have the same resolution. Therefore, we feed the whole image to \cmrnet2 without any cropping. However, we still downsample the images of the Argoverse dataset to have a similar field of view as the images used for training. Once we have the set of 2D-3D correspondences predicted by the network, we apply \ac{PnP}+RANSAC to estimate the pose of the camera \wrt the map, as detailed in \Cref{sec:localization}. We repeat this whole process three times using three specialized instances of \cmrnet2 to iteratively refine the estimation.
The inference time for a single iteration, on one NVIDIA GTX 1080Ti is about \SI{0.05}{\second} for the \ac{CNN} and \SI{1.25}{\second} for \ac{PnP}+\ac{RANSAC}.

\subsection{Results}
We evaluate the localization performance of \cmrnet2 on the validation sets of the three datasets described in \Cref{sec:datasets}. It is important to note that all the validation sets are geographically separated from the training set; thus, we evaluate \cmrnet2 in places that were never seen during the training phase. Moreover, as the Lyft5 dataset was not used for training, we evaluate the performance of our approach in a completely different city and on data gathered with different sensors. The median localization errors for the three iterations of the iterative refinement technique are reported in \Cref{tab:iterative}. 
Furthermore, videos showing qualitative results are available at \href{https://rl.uni-freiburg.de/research/vloc-in-lidar
}{\color{red}http://rl.uni-freiburg.de/research/vloc-in-lidar}.
We also provide comparisons of our proposed \cmrnet2 with CMRNet and the approach of~\cite{Caselitz_2016} in \Cref{tab:comparison}. The results demonstrate that \cmrnet2 outperforms the other methods on the sequence 00 of the KITTI dataset. Although \cite{Caselitz_2016}~achieves a lower standard deviation for the translation component, it further exploits tracking and temporal filtering which our approach does not. Therefore, the variance of our method can be further lowered by incorporating such techniques.

As opposed to CMRNet, \cmrnet2 can fail to localize an image. This might particularly occur in the first iteration, when more than half of the predicted matches are incorrect, and therefore \ac{PnP}+RANSAC estimates an incorrect solution. We believe the cause to be in the matching of pixels on the ground plane: due to the uniform appearance of the road surface, it is nearly impossible to recognize the exact pixel that matches a specific 3D point. To identify such cases, we mark the samples as \textit{failed} when the estimated pose, after the first iteration, is farther than four meters from $H_{init}$. Failed samples are not fed into the second and third iterations. The percentage of failed samples is also reported in \Cref{tab:iterative}.
%

\section{Conclusions}\label{sec:conclusion}

In this paper we proposed \cmrnet2, a novel \ac{CNN}-based approach for monocular localization in LiDAR-maps.
We designed \cmrnet2 to be independent of both the map and camera intrinsics. 
To the best of our knowledge, \cmrnet2 is the first \ac{DNN}-based approach for pose regression that generalizes to new environments without any retraining. 

We demonstrated the performance of our approach on KITTI, Agroverse, and Lyft5 datasets in which \cmrnet2 localizes a single RGB image in unseen places with a median error as low as \SI{0.14}{\metre} and \ang{0.43}, outperforming other state-of-the-art approaches. Moreover, results on the Lyft5 dataset, which was excluded from training, show that \cmrnet2 is able to effectively generalize to previously unseen environments as well as to different sensors.
We plan to investigate whether the differentiable RANSAC approach~\cite{Brachmann_2018_CVPR} could be employed to train \cmrnet2 in and end-to-end fashion, while maintaining the camera parameters outside of the learning step.
\newpage
\section*{Acknowledgments}
This work was partly funded by the European Union's Horizon 2020 research and innovation program under grant agreement No 871449-OpenDR, by the Federal Ministry of Education and Research (BMBF) of Germany under SORTIE. The authors would like to thank Pietro Colombo for assistance in making the video.

\balance




\bibliographystyle{IEEEtran}
\bibliography{IEEEabrv,bibliography}

\end{document}